\documentclass[11pt]{article}
\usepackage[T1]{fontenc}
\usepackage[utf8]{inputenc}
\usepackage{lmodern,microtype,amsmath,amssymb,booktabs,graphicx,array,xcolor}
\usepackage[margin=1in]{geometry}
\usepackage[colorlinks=true,urlcolor=blue,citecolor=blue,linkcolor=blue]{hyperref}
\usepackage{enumitem}
\usepackage{tikz}
\usetikzlibrary{positioning,arrows.meta,calc,decorations.pathreplacing}
\definecolor{tokG}{RGB}{214,214,214}\definecolor{tokP}{RGB}{190,214,238}
\definecolor{tokE}{RGB}{242,192,188}\definecolor{tokC}{RGB}{252,212,160}
\definecolor{tokR}{RGB}{214,202,232}\definecolor{tokO}{RGB}{190,228,216}
\definecolor{tokM}{RGB}{244,244,244}\definecolor{encF}{RGB}{232,243,224}
\definecolor{encT}{RGB}{206,229,192}\definecolor{encS}{RGB}{176,210,158}
\definecolor{headC}{RGB}{253,231,210}
\setlist{nosep}
\newcommand{\incr}[1]{$\uparrow$\,#1}
\newcommand{\decr}[1]{$\downarrow$\,#1}
\hypersetup{pdftitle={Zero-Search Reinforcement Learning for Slay the Spire 2 with Failure-Driven Restarts},
  pdfauthor={Xiangqin Chen, Zhengxuan Xie}}
\title{Zero-Search Reinforcement Learning for Slay the Spire 2\\with Failure-Driven Restarts}
\author{Xiangqin Chen\\University of Waterloo\\\texttt{xiangqin.chen@uwaterloo.ca}
\and Zhengxuan Xie\\University of Waterloo\\\texttt{z94xie@uwaterloo.ca}}
\date{November 2026}

\begin{document}
\maketitle

\begin{abstract}
Slay the Spire~2 (STS2) is a deck-building roguelike in which a winning run at the highest
difficulty takes several hundred decisions, and most losses happen in late boss fights that a
learner seldom reaches. We train agents that choose every action with one forward pass of a
6.6M-parameter entity transformer, without lookahead. Training and evaluation run
the game's own compiled assembly (v0.107.1) in a headless .NET host. The host replays a run exactly
from its seed and actions, so training episodes can restart from states the learner has visited, and
a failure-driven curriculum allocates these restarts in proportion to where complete runs are lost.
On 12,800 held-out games at Ascension~10 (A10), the released policies win 8.05\% averaged over the
five characters (Ironclad 15.6\%, Regent 12.5\%), and AlphaSpire's released networks used without
search win 0.55\% on a common panel. Longer A10 training raised Ironclad from 12.4\% to 15.6\% and
Defect from 3.0\% to 8.1\% on these games. At
Ascension~0 the released policies win 80.2\%, level with the 80.3\% that the tree-search agent Spire
Mentor reports, while Mentor leads by 5 to 8 points from A4 upward. Run in our host on the same 2,000
A10 games, Mentor's search wins 8.8\% against 8.2\% for our policies, which need 21--27~ms per decision
on one CPU core. GPT-6 Astra at maximum reasoning won none of 11 A10 games through our text interface.
In a controlled study of 18 training runs from six random
initializations, failure-driven allocation exceeds uniform restart allocation on each of three Regent
initializations, by 3.6 percentage points on average, while the Silent effect is small and changes
sign. Code, trained policies, an in-game mod and the records behind every table are at
\url{https://github.com/HalfVulpes/sts2ai}.
\end{abstract}

\section{Introduction}
A run of Slay the Spire~2 crosses three acts of procedurally generated maps. Each floor brings a
combat, event, shop, treasure or rest site, and the player builds a deck, collects relics and
potions, and chooses a route along the way. Ascension levels add cumulative modifiers, and the
highest level in the current version, Ascension~10 (A10), ends the run with two consecutive boss
fights. A deck assembled poorly in act one can make the final fights unwinnable, so combat skill
alone does not secure a win. The decisions that settle a run are also rare in the training data.
In an early development measurement at Ascension~0, the act-3 boss accounted for 7\% of an
Ironclad policy's decisions and 59\% of its losses.

Current STS2 agents reach their strongest results with large computation at decision time.
Spire Mentor combines a learned run-level policy with thousands of tree-search simulations in
every combat decision~\cite{mentor}.
We ask how far a learned policy goes when it chooses each action with a single network
evaluation, and which training changes help at the late bottleneck. A policy of this kind runs
inside the game on a CPU, and with the simulator included it plays a winning A10 game in about
30~seconds on two CPU cores.

Our approach rests on the game's own code. A headless host loads the installed game assembly and
exposes every in-game choice as a decision, and it reproduces a run exactly from its seed and the
sequence of chosen actions. This determinism lets a worker save the action prefix that led to an
intermediate state and later restart a training episode from it. We use this to give the learner
practice where it fails. A failure map tracks where complete runs from the ordinary start are lost,
and restart episodes are allocated across twelve situations (act start and the three kinds of
combat in each act) in proportion to that map.

This paper makes four contributions.
\begin{enumerate}
\item A training and evaluation platform that runs the STS2 v0.107.1 assembly headlessly, with a
  structured observation encoder shared by the trainer and an in-game mod, exact replay of runs,
  and an evaluation protocol on the game's canonical seed strings (Section~\ref{sec:env}).
\item Zero-search policies for all five characters, evaluated on 12,800 held-out A10 games and at
  four lower difficulties, with the code, the policies and a playable mod released
  (Sections~\ref{sec:agent} and~\ref{sec:released}).
\item A comparison of zero-search policies with Monte Carlo tree search and language models on the same
  game version, including Spire Mentor run in our host on identical seeds and GPT-6 Astra through a
  headless text interface (Section~\ref{sec:compare}).
\item A controlled 18-run study of failure-driven restart allocation from independent random
  initializations, which finds a consistent gain for one character and none for the other
  (Section~\ref{sec:screen}).
\end{enumerate}

\section{Related work}
\paragraph{Restart curricula.}
Starting episodes from intermediate states is an established way to help reinforcement learning
with long horizons. Reverse Curriculum Generation grows a set of start states outward from the
goal~\cite{florensa2017}, and Backplay starts episodes at late states of a demonstration and moves
the start point backward during training~\cite{resnick2018}. Go-Explore returns to promising archived
states before exploring further~\cite{ecoffet2021}. Jump-Start RL lets a guide policy produce the
first part of each episode for an exploration policy~\cite{uchendu2023}. Prioritized Level Replay
revisits training levels in proportion to an estimate of their learning potential~\cite{jiang2021},
and Per-Context Reverse Curricula adjusts the amount of assistance separately for each context from
success feedback~\cite{aphale2026}. Our restart states come from the learner's own past trajectories,
including failed ones, so a later state need not be an easier start. The novel element is the outer
allocation rule that spreads restarts across situations by where complete runs fail. Our controlled
study isolates this rule. It does not compare against faithful reimplementations of the methods above.

\paragraph{Agents for Slay the Spire.}
For STS2, AlphaSpire releases a policy-and-search system with networks for run-level and combat
decisions that guide a tree search~\cite{alphaspire}. Spire Mentor trains a run-level policy with
PPO and plays combat with Monte Carlo tree search at several thousand simulations per decision; it
reports win rates for all five characters at five ascension levels~\cite{mentor}. The
sts2-rl-research project trains separate combat and choice agents without search and reports
26.6\% for Ironclad at Ascension~2~\cite{shuchilab}. Epoch AI's SpireClimb evaluates frontier
language models on STS2 at A10 from screenshots with access to the game wiki~\cite{spireclimb}. Its
published data do not link per-run logs, so its individual runs cannot be checked. In 8 Defect games
that we ran with the same model in a similar setting, Astra won none, but we recorded neither their
seeds nor their screenshots (Section~\ref{sec:compare}). AgenticSTS studies bounded memory for
language-model agents on STS2, mainly with Silent at Ascension~0~\cite{agenticsts}. SpireBench
evaluates language models on the original Slay the Spire at Ascension~20~\cite{spirebench}.

\paragraph{Evaluation.}
We follow recommendations to report uncertainty over independent training runs separately from
uncertainty over evaluation episodes, and to fix evaluation protocols before
testing~\cite{agarwal2021,patterson2023}.

\section{Environment and evaluation protocol}
\label{sec:env}
\subsection{Headless host}
STS2 is a Godot/.NET application. The host loads the installed game assembly into a .NET~9 process
with a minimal Godot compatibility layer and the game's test mode. It advances the asynchronous
game logic on a deterministic synchronization context and requests decisions through the entry
points the user interface calls. Patches remove input/output side effects, restore real-game
behavior where test mode takes shortcuts, and make accesses to interface nodes null-safe; a static
rewriter covers about 500 methods that touch interface nodes. Any exception raised by the game fails
the episode, and evaluations report such failures. The host advances prompts with exactly one
option and gold rewards automatically, and plays the Crystal Sphere minigame with a fixed local
heuristic. The policy, the text interface and every comparison in this paper share these
conventions. The research version is the Linux build of v0.107.1 with all content unlocked; the
repository records the hashes of all eleven game assemblies, and an importer verifies them on a
user's own installation.

A run is reproduced from its seed and the sequence of chosen action indices. Restart training relies
on this. Before a restarted episode continues, the worker compares a fingerprint of the encoded
observation at the reconstructed state with the stored fingerprint and treats a mismatch as an
error. In the final calibration run, 3,765 restart episodes completed without a mismatch. Because
the fingerprint covers only the encoded observation, hidden game variables are checked only
indirectly, through later observations.

\subsection{Observation and actions}
Each decision is encoded as up to 192 typed entity tokens: a global run-state token, the player,
enemies with their intents, powers, cards in hand and the play pile, multisets of the draw, discard,
exhaust and deck piles, orbs, potions, relics, the act bosses, offers such as shop items, rewards,
event options and map nodes, and a summary of the run history. A token carries five categorical
identifiers and 32 numeric features. Draw piles are encoded as multisets without order, and the
observation contains no random-number state. The encoding includes the internal identifier of each
enemy's next move as well as its rendered intent, so the observation contains slightly more than a
human sees on screen. Each legal action is a token with its own features and pointers to up
to two entity tokens, such as a card and its target. Typically 5 to 40 actions are legal, at most 128.

\subsection{Seeds, panels and statistics}
The game turns a ten-character seed string from a base-34 alphabet without I and O into its random
streams. Our evaluations use these canonical strings and the game's own hashing of the act-selection
stream, so a seed means the same in our host as in the native game's seed entry. Earlier
development evaluations used a different numeric mapping; we keep them separate and label them as
legacy where they appear. Every evaluation panel is drawn uniformly at random with a recorded
generator seed before the evaluated policies are trained or chosen. Training samples seeds uniformly
from the same range, excluding every evaluation panel; a panel used to choose a checkpoint is never
reused to report that checkpoint's result.

We report win rates with Wilson 95\% intervals, which describe sampling variation of a fixed policy
over games. Comparisons of two policies on the same seeds use the exact McNemar test on the
discordant games, with a Bonferroni or Holm correction stated for each family of comparisons. For
training methods, the unit of replication is the independent training run, and game-level intervals
do not measure variation between training runs.

\section{Agent}
\label{sec:agent}
\subsection{Network}
Token embeddings sum the five identifier embeddings with a projection of the numeric features, and
eight learned memory tokens are appended. Six pre-norm transformer blocks (width 256, eight heads,
feed-forward width 1024) encode the set. The state summary concatenates the output of the global
token with attention pooling over all tokens. The policy scores each legal action with a three-layer
MLP on the action embedding, a projection of the entity tokens it points to, and the state summary,
followed by a softmax over legal actions. A value trunk feeds four heads: discounted return,
probability of winning the run, floors remaining, and end-of-combat HP fraction; the last three are
auxiliary targets. With the vocabulary of 2,650 entries the network has 6,568,453 parameters. At
evaluation the agent takes the highest-scoring legal action of the exponential moving average (EMA)
of the weights. The same encoder and network run in C\# inside the game as a mod with autoplay,
single-stepping and an overlay of the policy.

\begin{figure}[ht]
\centering
\begin{tikzpicture}[font=\footnotesize,>={Stealth[length=1.6mm,width=1.3mm]},
  pill/.style={draw=black!50,line width=0.4pt,rounded corners=1.2pt,minimum width=12.5mm,minimum height=3.4mm,inner sep=0pt,font=\scriptsize},
  cell/.style={draw=black!45,line width=0.4pt,minimum size=3.4mm,inner sep=0pt},
  set/.style={draw=black!30,fill=black!2,rounded corners=3pt,line width=0.4pt},
  box/.style={draw=black!60,align=center,inner sep=2pt,line width=0.5pt},
  sub/.style={box,rounded corners=1pt,fill=white,minimum width=22mm,minimum height=4.2mm,font=\scriptsize},
  plus/.style={circle,draw=black!60,fill=white,inner sep=0.4pt,font=\scriptsize},
  flow/.style={->,line width=0.6pt,black!70},
  ptr/.style={->,densely dashed,line width=0.5pt,black!55},
  lab/.style={font=\scriptsize,align=center}]
\draw[set] (0.48,0.85) rectangle (1.92,5.65);
\foreach \y/\c/\t in {5.40/tokG/global,4.98/tokP/player,4.56/tokE/enemy,4.14/tokC/card,3.72/tokC/card,3.30/tokR/relic,2.88/tokO/offer}
  \node[pill,fill=\c] at (1.2,\y) {\t};
\node at (1.2,2.47) {$\vdots$};
\foreach \y in {1.94,1.52,1.10} \node[pill,fill=tokM] at (1.2,\y) {memory};
\draw[decorate,decoration={brace,amplitude=3pt,mirror},line width=0.4pt] (0.38,5.57) -- (0.38,2.30);
\node[lab,rotate=90] at (0.14,3.935) {entity tokens ($\le$192)};
\draw[decorate,decoration={brace,amplitude=3pt,mirror},line width=0.4pt] (0.38,2.11) -- (0.38,0.93);
\node[lab,rotate=90] at (0.14,1.52) {8 memory};
\node[lab] at (1.2,0.42) {5 identifiers and\\32 numeric features};
\draw[flow] (1.92,3.25) -- node[above,lab]{embed} (2.85,3.25);
\foreach \k in {5,4,3,2,1,0} {
  \pgfmathsetmacro{\d}{0.12*\k}
  \filldraw[fill=encS,draw=black!45,line width=0.3pt] (5.75+\d,0.75+\d) -- ++(0.12,0.12) -- ++(0,5.0) -- ++(-0.12,-0.12) -- cycle;
  \filldraw[fill=encT,draw=black!45,line width=0.3pt] (2.85+\d,5.75+\d) -- ++(0.12,0.12) -- ++(2.9,0) -- ++(-0.12,-0.12) -- cycle;
  \filldraw[fill=encF,draw=black!55,line width=0.4pt] (2.85+\d,0.75+\d) rectangle ++(2.9,5.0);
}
\node[sub] (ln1) at (4.15,1.25) {LayerNorm};
\node[sub,minimum height=7.6mm] (att) at (4.15,2.05) {self-attention\\8 heads, $d{=}256$};
\node[plus] (p1) at (4.15,2.80) {$+$};
\node[sub] (ln2) at (4.15,3.40) {LayerNorm};
\node[sub,minimum height=7.6mm] (ffn) at (4.15,4.20) {feed-forward\\$256{\to}1024{\to}256$};
\node[plus] (p2) at (4.15,4.95) {$+$};
\draw[flow] (4.15,0.75) -- (ln1); \draw[flow] (ln1) -- (att); \draw[flow] (att) -- (p1);
\draw[flow] (p1) -- (ln2); \draw[flow] (ln2) -- (ffn); \draw[flow] (ffn) -- (p2); \draw[flow] (p2) -- (4.15,5.75);
\draw[flow,rounded corners=2pt] (4.15,0.92) -| (5.48,2.80) -- (p1);
\draw[flow,rounded corners=2pt] (4.15,3.08) -| (5.48,4.95) -- (p2);
\node[lab] at (4.3,0.3) {transformer encoder\\6 pre-norm blocks};
\draw[flow] (6.52,3.25) -- (7.17,3.25);
\draw[set] (7.17,0.85) rectangle (7.73,5.65);
\foreach \y/\c [count=\i] in {5.40/tokG,4.98/tokP,4.56/tokE,4.14/tokC,3.72/tokC,3.30/tokR,2.88/tokO}
  \node[cell,fill=\c!60] (o\i) at (7.45,\y) {};
\node at (7.45,2.47) {$\vdots$};
\foreach \y in {1.94,1.52,1.10} \node[cell,fill=tokM] at (7.45,\y) {};
\node[lab] at (7.45,0.6) {outputs $h_i$};
\node[box,fill=encT!50,minimum width=24mm] (sum) at (9.45,5.85) {state summary $s$\\{\scriptsize global token $\oplus$}\\{\scriptsize attention pool}};
\draw[flow] (7.45,5.65) |- (sum.west);
\node[box,fill=headC,minimum width=13mm,minimum height=8mm] (val) at (11.75,5.85) {value\\trunk};
\draw[flow] (sum.east) -- (val.west);
\coordinate (bus) at (12.62,0);
\foreach \y/\t [count=\i] in {6.48/{return $V(s)$},6.06/{$P(\mathrm{win})$},5.64/{floors left},5.22/{HP fraction}}
  \node[pill,fill=headC!60,minimum width=20mm] (h\i) at (14.0,\y) {\t};
\draw[line width=0.6pt,black!70] (val.east) -- (val.east -| bus);
\draw[line width=0.6pt,black!70] (h1.west -| bus) -- (h4.west -| bus);
\foreach \i in {1,...,4} \draw[flow] (h\i.west -| bus) -- (h\i.west);
\draw[decorate,decoration={brace,amplitude=3pt},line width=0.4pt] (15.12,6.23) -- (15.12,5.05);
\node[lab,rotate=90] at (15.42,5.64) {auxiliary};
\foreach \y/\n/\t in {3.02/a1/a_1,2.60/a2/a_2,1.80/ak/a_k}
  \node[pill,fill=white,minimum width=5.5mm] (\n) at (9.0,\y) {$\t$};
\node at (9.0,2.24) {$\vdots$};
\draw[ptr] (o3.east) to[out=0,in=180] (a1.west);
\draw[ptr] (o4.east) to[out=0,in=180] (a1.west);
\draw[ptr] (o5.east) to[out=0,in=180] (a2.west);
\node[lab,anchor=west,align=left,black!60] at (8.1,4.68) {pointers\\($\le$2 per action)};
\node[lab] at (9.0,1.2) {legal actions\\($\le$128)};
\draw[decorate,decoration={brace,amplitude=3pt},line width=0.4pt] (9.38,3.19) -- (9.38,1.63);
\foreach \o in {0.18,0.09}
  \filldraw[fill=headC!88!black,draw=black!50,line width=0.4pt] (10.05+\o,1.95+\o) rectangle ++(2.6,0.9);
\node[box,fill=headC,minimum width=26mm,minimum height=9mm] (pol) at (11.35,2.40) {policy MLP\\{\scriptsize 3 layers on $[a_j;\,s]$}};
\draw[flow] (9.55,2.41) -- (pol.west);
\coordinate (sx) at (10.52,0);
\draw[flow] (sum.south -| sx) -- node[pos=0.72,left,lab]{$s$} (10.52,3.03);
\draw[flow] (12.83,2.40) -- node[above,lab]{logits} (13.75,2.40);
\draw[line width=0.4pt,black!60] (13.85,1.95) -- (15.25,1.95);
\foreach \x/\h in {14.00/0.22,14.27/0.80,14.54/0.12,14.81/0.38,15.08/0.08}
  \filldraw[fill=headC!80!orange,draw=black!50,line width=0.3pt] (\x-0.09,1.95) rectangle (\x+0.09,1.95+\h);
\node[lab] at (14.55,3.0) {$\pi(a_j\mid s)$};
\node[lab] at (14.55,1.5) {softmax over\\legal actions};
\end{tikzpicture}
\caption{Policy and value network (6,568,453 parameters). Typed entity tokens and eight learned memory
tokens pass through six pre-norm transformer blocks; the front block shows their layers. The value
trunk reads a state summary of the encoder outputs. The same MLP scores every legal action from its
own features, the encoder outputs of the entities it points to (dashed) and the state summary.}
\label{fig:arch}
\end{figure}
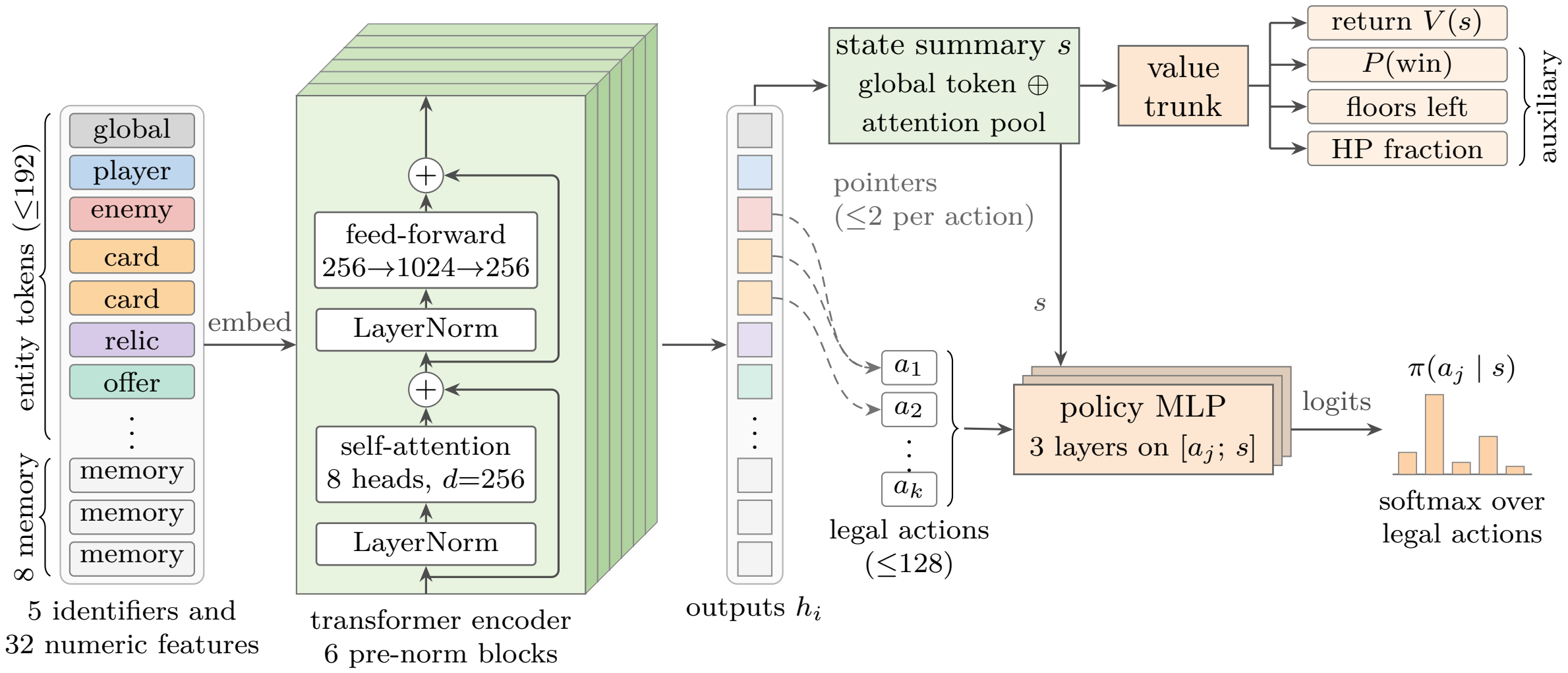

\subsection{Training}
We train with clipped PPO~\cite{schulman2017}, using separate entropy coefficients for combat and
non-combat decisions, $\gamma=0.9998$, GAE $\lambda=0.99$, 65,536 samples per iteration, two epochs
and minibatches of 4,096. The reward gives $+0.01$ per floor, $+0.02$ per elite, $+0.1$ per boss,
$+1$ for a win and $-0.1$ for a death, plus a potential-based term on the HP fraction. We make no
claim that this shaped objective has the same optimum as the win rate. Episodes end as losses after
30 identical non-combat actions, 150 combat turns or 300 actions in a turn, which removes reward
loops. A native collector multiplexes several hundred simulator processes through shared memory and
batches their inference on the GPU.

The policies start from an exploratory development history on two GPU servers. Training began with
all characters sharing one network, continued with single-character specialists, an ascension
curriculum and the restart curriculum of Section~\ref{sec:method}, and finished with a learning rate
of $10^{-4}$ and training directly at A10 for Ironclad and Regent. We call the final checkpoints of
that history the base policies. They are EMA weights after 0.93B (Defect), 1.14B (Silent), 1.17B
(Necrobinder), 1.66B (Ironclad) and 1.68B (Regent) collected decisions, each matches a historical EMA
tensor exactly, and they were chosen on legacy development seeds. The released Regent and Necrobinder
policies are their base checkpoints. The released Ironclad, Defect and Silent policies continued from
their base weights for 1,800 updates (118M collected decisions) at A10 with the failure-driven
curriculum and a learning rate of $10^{-4}$, under a protocol frozen before training that named each
run's final EMA as its only candidate (Section~\ref{sec:improve}). Every canonical panel in this paper
was drawn after the base checkpoints were chosen and is excluded from the training seeds of these runs.

\section{Failure-driven restart curriculum}
\label{sec:method}
Let $\rho_0$ be the distribution of ordinary run starts. The objective is the win rate from these
starts,
\begin{equation}
J_0(\theta)=\mathbb{E}_{s_0\sim\rho_0,\,\pi_\theta}\left[\mathbf{1}\{\text{run won}\}\right].
\end{equation}
Restart episodes begin at intermediate states and therefore change the distribution of training
starts. The PPO probability ratio corrects for the change of policy and leaves this change
uncorrected, so a restart update is a heuristic step for $J_0$. We always evaluate on complete runs
from $\rho_0$.

\paragraph{Situations.} Restart states fall into twelve situations: the start of each act, and the
first decision of a monster, elite or boss combat in each act. Each worker keeps up to 16 recent
states per situation in a first-in-first-out buffer. A state is stored as the game seed, character,
ascension and the action prefix that reaches it.

\paragraph{Failure map.} Every ten PPO iterations, the losses of fresh single-player runs since the
last update form an attribution vector $D_t\in\mathbb{R}^{12}_{\ge 0}$. A loss in combat credits
0.65 to that combat's situation and 0.35 to the start of its act, which stands for the deck-building
choices made since then. A loss outside combat credits the start of its act. The map and the
situation distribution are
\begin{align}
L_t &= 0.5\,L_{t-1}+D_t,\\
p_t(s) &= \frac{L_t(s)/\sum_j L_t(j)+0.02}{1+12\cdot 0.02}.
\end{align}
The distribution starts uniform and keeps its previous value when an update has no new losses. It
measures where losses occur, and it is not normalized by how often a situation is visited.

\paragraph{Episode resets.} At each reset a worker starts a restart episode with probability 0.5.
It samples a situation from $p_t$ restricted to the situations for which it holds a stored state,
\begin{equation}
p_t(s\mid A_w)=\frac{\mathbf{1}\{s\in A_w\}\,p_t(s)}{\sum_{j\in A_w}p_t(j)},
\end{equation}
which keeps the actual restart rate equal to the requested rate when some buffers are empty.
Within the situation, a state with $n_i$ restarts and $\ell_i$ losses is sampled with weight
\begin{equation}
w_i=0.1+4q_i(1-q_i),\qquad q_i=\frac{\ell_i+1}{n_i+2},
\end{equation}
which favors states whose outcome is uncertain. The worker replays the prefix without querying the
learner and then plays on with the current policy. Replayed decisions cost simulator time and do not
enter the PPO update. Figure~\ref{fig:proc} summarizes the procedure.

\begin{figure}[ht]
\centering
\fbox{\begin{minipage}{0.93\linewidth}\small
\textbf{Failure-driven restart curriculum} (one learner, $W$ workers)\par\smallskip
\textit{Worker $w$, at every episode reset:}
\begin{enumerate}[leftmargin=2em]
\item With probability $0.5$ and a nonempty set $A_w$ of situations with stored states, draw
  $s\sim p_t(\cdot\mid A_w)$, draw a stored state $i$ of $s$ with weight $w_i$, replay its action
  prefix from its seed, and check the observation fingerprint. Otherwise start a fresh run.
\item Play with the current policy. On entering any of the 12 situations in a fresh run, store
  (seed, action prefix, fingerprint) in that situation's 16-slot buffer.
\item At the end of a restart episode, update $n_i$ and $\ell_i$ of the state it started from.
\end{enumerate}
\textit{Learner, every PPO iteration:} update on the collected decisions (replayed prefixes
excluded). Every ten iterations, attribute the losses of fresh runs to situations ($D_t$), update
$L_t$ and $p_t$, and send $p_t$ to the workers.
\end{minipage}}
\caption{The restart procedure. The uniform control replaces $p_t$ with the uniform distribution
over the same 12 situations and keeps everything else.}
\label{fig:proc}
\end{figure}

\paragraph{Losses are a heuristic for value.} Consider two necessary stages that succeed
independently with $p_1=0.6$ and $p_2=0.5$, so $J_0=0.3$. Losses fall on the first stage with mass
$0.4$ and on the second with mass $0.6\cdot0.5=0.3$, so the failure map favors stage one. If training
raises either stage's success probability by the same amount, the gain in $J_0$ is
$\partial J_0/\partial p_1=p_2=0.5$ for stage one and $\partial J_0/\partial p_2=p_1=0.6$ for
stage two, and stage two is the better use of practice. Loss counts can therefore point practice
to the wrong stage, and the benefit of the rule has to be measured.

\section{Released policies}
\label{sec:released}
Table~\ref{tab:a10} pools three disjoint held-out A10 panels: the 512-game panel of the external
comparison (Section~\ref{sec:compare}) and the two 1,024-game test panels of the selection and
continuation studies (Section~\ref{sec:improve}), where the base policies serve as the baseline. All
12,800 games of the released policies completed with no engine error. Ironclad wins about one run in
six, Regent one in eight and Defect one in twelve; Silent and Necrobinder win about 2\%. On these games
the continued Ironclad and Defect policies win clearly more often than their base policies (269
against 187 seeds won by only one of the two, exact McNemar $p<10^{-3}$, and 186 against 56,
$p<10^{-16}$), while Silent's rise from 47 to 61 wins is not significant (54 against 40, $p=.18$). The
five-character mean rises from 6.29\% to 8.05\%.

\begin{table}[ht]
\centering\small
\caption{Released zero-search policies at A10 on 2,560 held-out canonical games per character
(three disjoint panels of 512, 1,024 and 1,024 games). Base: the policy before the release
continuation, on the same games; Regent and Necrobinder are released unchanged. Bold marks the higher
count, arrows the change in win rate; paired exact McNemar $p$. Intervals are Wilson 95\%.}
\label{tab:a10}
\begin{tabular}{lrrrrrr}
\toprule
Character & Base & Released & Win rate (\%) & 95\% interval (\%) & Change (pp) & $p$\\
\midrule
Ironclad & 317 & \textbf{399} & 15.59 & 14.23--17.04 & \incr{3.20} & $<10^{-3}$\\
Silent & 47 & \textbf{61} & 2.38 & 1.86--3.05 & \incr{0.55} & .18\\
Defect & 78 & \textbf{208} & 8.12 & 7.13--9.25 & \incr{5.08} & $<10^{-16}$\\
Necrobinder & \multicolumn{2}{r}{42 (unchanged)} & 1.64 & 1.22--2.21 & & \\
Regent & \multicolumn{2}{r}{321 (unchanged)} & 12.54 & 11.31--13.88 & & \\
\midrule
Mean of five & 805 & \textbf{1031} & 8.05 & & \incr{1.77} & \\
\bottomrule
\end{tabular}
\end{table}

Table~\ref{tab:sweep} reports the same policies at Ascension 0, 4, 8 and 9 on 512 common seeds per
level, which match the levels that Spire Mentor reports. All 10,240 games completed without engine
error. The five-character mean falls from 80.2\% at A0 to 71.5\% at A4, 51.8\% at A8 and 22.8\% at
A9, then to 8.1\% at A10. Ironclad and Regent keep about two fifths of their A9 rate at A10 (39.3\% to
15.6\% and 30.1\% to 12.5\%), where the run continues into a second final boss. Silent and
Necrobinder are already below 13\% at A9. Against their base policies on the same seeds, the
continued Defect policy wins more often at A4, A8 and A9 (all $p<.01$), the continued Ironclad policy
wins more often at A8 (370 against 329 of 512, $p=.002$) and less often at A0 (460 against 479,
$p=.027$), and Silent changes significantly at no level.

\begin{table}[ht]
\centering
\caption{Released zero-search policies across difficulty, win rate in percent. A0--A9: 512 common
canonical seeds per level; A10: the 2,560 games per character of Table~\ref{tab:a10}. Bold marks the
best character at each level; arrows mark a significant change of a continued policy against its base
policy on the same seeds (exact McNemar, $p<.05$).}
\label{tab:sweep}
\begin{tabular}{lrrrrr}
\toprule
Character & A0 & A4 & A8 & A9 & A10\\
\midrule
Ironclad & \textbf{89.8}\,$\downarrow$ & \textbf{88.3} & \textbf{72.3}\,$\uparrow$ & \textbf{39.3} & \textbf{15.6}\,$\uparrow$\\
Silent & 67.6 & 54.3 & 36.1 & 12.3 & 2.4\\
Defect & 83.8 & 77.0\,$\uparrow$ & 60.0\,$\uparrow$ & 26.6\,$\uparrow$ & 8.1\,$\uparrow$\\
Necrobinder & 72.7 & 59.4 & 34.2 & 5.9 & 1.6\\
Regent & 87.1 & 78.5 & 56.4 & 30.1 & 12.5\\
\midrule
Mean of five & 80.2 & 71.5 & 51.8 & 22.8 & 8.1\\
\bottomrule
\end{tabular}
\end{table}

A10 adds a second final boss directly after the first. In a development sweep of an earlier
Ironclad policy on 100 legacy-protocol seeds, the second final boss ended 25 of the 100 A10 runs,
the largest single cause, and the act-1 boss ended 19. This failure profile motivated training
directly at A10 (Section~\ref{sec:improve}).

\section{Comparison across agent classes}
\label{sec:compare}
Table~\ref{tab:compare} places the released policies next to the agents we ran ourselves and the
zero-search and tree-search STS2 A10 results reported by others, grouped by the computation they spend
on a decision. The agents also differ in what they observe.

\begin{table}[ht]
\centering\small
\caption{STS2 v0.107.1, Ascension 10, all five characters. ``Measured'' rows are our runs; ``reported''
rows are taken from the cited sources, with their own protocols and seed panels.}
\label{tab:compare}
\begin{tabular}{p{0.24\linewidth}p{0.29\linewidth}lrr}
\toprule
Agent & Computation per decision & Source & Games & Mean win (\%)\\
\midrule
Ours (released) & one network evaluation, 21--27 ms on one CPU core & measured & 12{,}800 & \textbf{8.05}\\
AlphaSpire networks without search~\cite{alphaspire} & one network evaluation & measured & 2{,}560 & 0.55\\
Spire Mentor~\cite{mentor} & 8{,}000 MCTS simulations per combat decision, about 112 ms on one core & reported & 2{,}500 & 14.0\\
Spire Mentor, in our host & 4{,}000--16{,}000 (Ironclad) or 8{,}000 simulations per combat decision & measured & 2{,}000 & 8.8\\
GPT-6 Astra, our text interface & maximum reasoning, median 13.7 s & measured & 11 & 0.0\\
DeepSeek V4.1 Flash 1017, our text interface & maximum reasoning, median 29 s & measured & 12 & 0.0\\
\bottomrule
\end{tabular}
\end{table}

\paragraph{Zero-search policies.} We ran AlphaSpire v0.2.1 with its released checkpoints in its
native greedy mode for both combat and run-level decisions, and required its counters of searched
decisions and search steps to stay at zero in every game. On a common panel of 512 seeds per
character, our policies win 208 games and AlphaSpire's networks win 14, all for Ironclad
(Table~\ref{tab:external}). The two systems run in their own engines: AlphaSpire uses its own
simulator in its custom mode, and our host uses Standard mode with all content unlocked. AlphaSpire's
checkpoints are trained as priors for search, and the non-Ironclad networks often end turns early
when used alone. In 61 Ironclad games, including 4 wins, AlphaSpire's action head fell back to a
uniform prior 227 times when the number of actions exceeded its axis. All these games are counted.
The table therefore compares two released artifacts used without search. It does not measure
AlphaSpire's search system. The sts2-rl-research agents also act without search, but they report
only Ironclad at Ascension~2 (26.6\%)~\cite{shuchilab}, so they do not enter the A10 table. In our
host, the released Ironclad policy wins 88.3\% at the harder Ascension~4.

\begin{table}[ht]
\centering
\caption{Released policies on a common panel of 512 canonical A10 seeds per character, both used
without search, wins of 512; bold marks the higher count. Wilson 95\% intervals in percent. All 5,120
games completed without engine error.}
\label{tab:external}
\begin{tabular}{lrrrr}
\toprule
Character & Ours & Interval & AlphaSpire & Interval\\
\midrule
Ironclad & \textbf{75} & 11.85--17.97 & 14 & 1.64--4.54\\
Silent & \textbf{9} & 0.93--3.31 & 0 & 0.00--0.74\\
Defect & \textbf{47} & 6.97--11.99 & 0 & 0.00--0.74\\
Necrobinder & \textbf{9} & 0.93--3.31 & 0 & 0.00--0.74\\
Regent & \textbf{68} & 10.61--16.50 & 0 & 0.00--0.74\\
\bottomrule
\end{tabular}
\end{table}

\begin{figure}[ht]
\centering
\includegraphics[width=\linewidth]{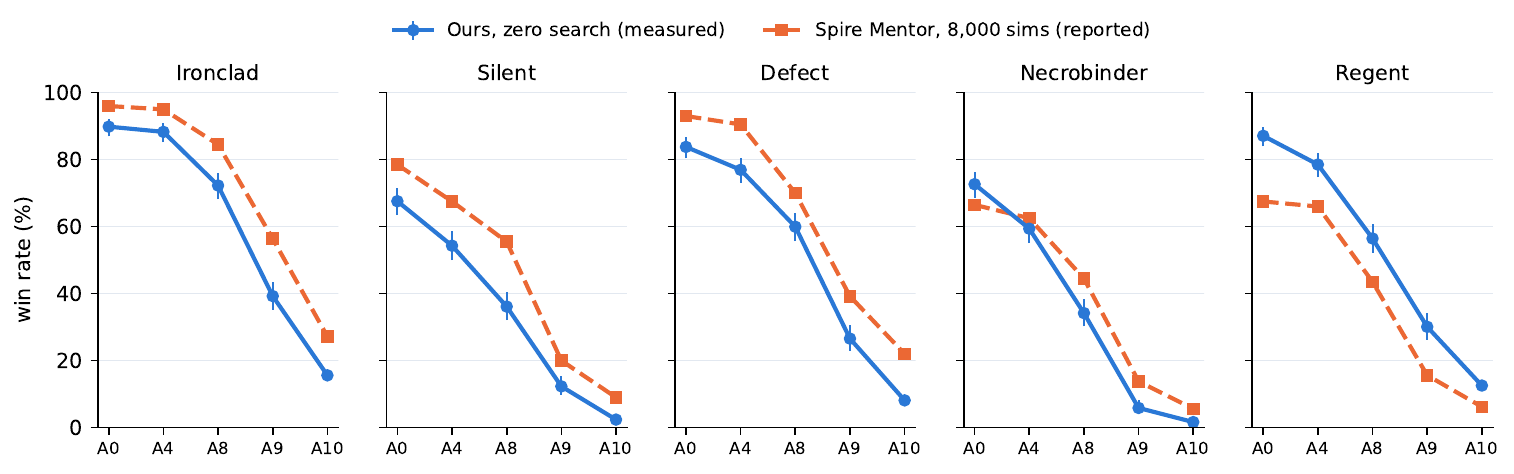}
\caption{Win rate by ascension. Our released zero-search policies (measured; 512 games per point at
A0--A9 and 2,560 at A10, Wilson 95\% intervals) and Spire Mentor at 8,000 simulations per combat
decision (reported; 200 runs per point at A0--A8 and 500 at A9--A10, no intervals given). Engines and
seed panels differ, and Mentor notes that its A10 values come from an engine version that favored A10.}
\label{fig:profile}
\end{figure}

\paragraph{Tree search.} Spire Mentor's rows use 8,000 simulations per combat decision with 500 runs
per character at A10, measured in October 2026~\cite{mentor}: Ironclad 27.2\%, Defect 22.0\%,
Silent 9.0\%, Regent 6.2\% and Necrobinder 5.6\%. The authors note that these rows were measured on
an engine version that granted rewards after the first of the two final bosses, a bug in A10's favor
that they later fixed. Their newer Ironclad policy wins 30.0\% of 400 runs on the corrected engine at
4,000 simulations per decision (16,000 in boss fights). On different engines and seed panels, our
zero-search Regent policy (12.5\%) wins more often than Mentor's searched Regent (6.2\%), and Mentor
leads on the other four characters, most clearly on Defect. Figure~\ref{fig:profile} extends the
comparison to the lower ascensions. At A0 the two agents are level on average (80.2\% against
80.3\%), and Mentor leads by 4.8 points at A4, 7.8 at A8, 6.2 at A9 and 5.9 at A10. Our Regent policy
leads Mentor's at every level, and our Necrobinder policy leads at A0.

\paragraph{Spire Mentor in our host.} To remove the engine and panel differences, we ran Spire Mentor
inside our host on 400 common canonical A10 seeds per character, the first 400 of a panel on which our
policies were also evaluated. Mentor's own released run-level networks make every non-combat
decision through its real-game adapter (map, card rewards, reward screens, shops, rest sites, events,
treasure and deck edits), and its search plays every combat move at its published budgets:
4,000 simulations per decision for Ironclad (16,000 in boss fights) and 8,000 for the other
characters. The search sees only what a player sees. Four details differ from Mentor's own real-game
driver: our policy makes the card choices that some cards open during combat (8 to 39 per run,
depending on the character), the search tree is rebuilt from the board at every decision, Mentor's
additional Ironclad search settings are not applied, and one rare event is skipped. All runs finished
without error, and no non-combat decision fell back to a random choice or to our policy.
Table~\ref{tab:mentorhost} gives the results. On identical games Mentor's search averages 8.8\% over
the five characters, against the 14.0\% it reports in its own engine, and our zero-search policies
average 8.2\%. Search leads clearly only on Silent (6.8\% against 2.3\%, $p=.004$). It also leads on
Ironclad (16.8\% against 13.3\%, $p=.13$) and Defect (12.5\% against 9.0\%, $p=.098$), but not
significantly; Necrobinder is level (3.5\% against 2.5\%, $p=.52$); and our Regent policy wins three
times as often as Mentor's (13.8\% against 4.5\%). Mentor spends three to five
minutes of search per run; our policy spends 21--27\,ms per decision.

\begin{table}[ht]
\centering\small
\caption{Spire Mentor (search) and our released policies (zero search) in the same host on the same
400 canonical A10 seeds per character. Wins of 400; bold marks the higher count; Wilson 95\% intervals
in percent; paired exact McNemar test on the seeds won by only one agent.}
\label{tab:mentorhost}
\begin{tabular}{lrrrrr}
\toprule
Character & Mentor & Interval & Ours & Interval & Only Mentor / only ours ($p$)\\
\midrule
Ironclad & \textbf{67} & 13.4--20.7 & 53 & 10.3--16.9 & 44 / 30 (.13)\\
Regent & 18 & 2.9--7.0 & \textbf{55} & 10.7--17.5 & 16 / 53 ($<$.001)\\
Silent & \textbf{27} & 4.7--9.6 & 9 & 1.2--4.2 & 27 / 9 (.004)\\
Defect & \textbf{50} & 9.6--16.1 & 36 & 6.6--12.2 & 38 / 24 (.098)\\
Necrobinder & \textbf{14} & 2.1--5.8 & 10 & 1.4--4.5 & 13 / 9 (.52)\\
\midrule
Mean of five & \textbf{8.8} & & 8.2 & & \\
\bottomrule
\end{tabular}
\end{table}

\paragraph{Swapping the two layers.} Both agents split a run into combat moves and everything else
(map, card rewards, shops, rest sites, events). On the same 400 seeds we crossed the two layers for
the base Ironclad and Regent policies, pairing our policy's non-combat decisions with Mentor's search
in combat, and the reverse (Table~\ref{tab:swap}). Each agent's non-combat play suits its own combat play. With our
policy in combat, Mentor's non-combat decisions collapse to 22 and 2 wins, while our policy's own win
41 and 55 (pooled, 90 against 18 seeds won by only one arm, $p<10^{-10}$). Mentor's search playing
the decks our policy builds is the strongest combination for Regent, with 70 wins against 55 for our
policy alone and 18 for Mentor alone. Over both characters, search on our decks gains 83 seeds and
loses 57 against our policy's own combat play ($p=.034$), about the three points that search at the
final bosses alone adds (Section~\ref{sec:improve}). Our policies' gap to search therefore lies in
combat; their run-level decisions hold up against Mentor's.

\begin{table}[ht]
\centering\small
\caption{Crossing non-combat decisions and combat play on the same 400 A10 seeds, wins of 400, for
the base Ironclad and Regent policies. The diagonal cells are the two agents unchanged; bold marks the
best combination per character.}
\label{tab:swap}
\begin{tabular}{llrr}
\toprule
Character & Non-combat decisions & Mentor's search in combat & Our policy in combat\\
\midrule
Ironclad & Mentor & \textbf{67} & 22\\
 & ours & 52 & 41\\
Regent & Mentor & 18 & 2\\
 & ours & \textbf{70} & 55\\
\bottomrule
\end{tabular}
\end{table}

\paragraph{Language models.} We ran GPT-6 Astra at its maximum reasoning setting on up to 25 A10 games,
five per character, drawn at random from the 512-seed panel of Table~\ref{tab:external} before any game
started. The model receives a structured text rendering of the player-visible state, including card,
relic and power descriptions, enemy intents and the unordered draw pile, together with numbered legal
actions. It has no tools, wiki or rollouts. Each request carries the current state, a strategic note of
at most 1,200 characters that the model maintains, and its four most recent action notes. States with a
single legal action advance automatically. Because of the cost, an amendment frozen before any game had
ended replaced the fixed 25 games with a Wald sequential test of $p=0.347$ against $p=0.10$ with error
rates of 5\% in both directions. Games are judged in their fixed launch order, using only the longest
prefix of finished games, since losses end earlier than wins. The test stops for the lower rate after
0 wins in 10 games, 1 in 15 or 2 in 19, and for the higher rate after 3 wins with at most 2 losses or 4
with at most 6. Astra lost all 11 games it played, including the first 10 in launch order, which ended
the cohort (one-sided 95\% upper bound on its win rate 24\%); a cost cap of ten games, set while the
last games were running, would have ended it at the same point. Astra played well into the run,
reaching the final bosses in 5 of 11 games and a mean floor of 34.5, against 32.8 for our released
policy on the same seeds, which won one of them. A game took a median of 395 model decisions and 2.2 hours of
model time; the cohort used 58.8M input and 2.55M output tokens. This result concerns our interface,
with bounded context, text instead of screenshots, and no wiki.

We also ran a smaller check from screenshots. On 8 November 2026, OpenAI's Codex agent with GPT-6 Astra
at its maximum reasoning setting opened the live game, the same version as above, on a separate virtual
machine and played Defect at A10 through computer use, with access to the game wiki and unrestricted
network access. It lost all 8 games, after which we stopped for cost (about 5,000 Codex credits per
game). These games had no frozen protocol. We did not record their seeds, and Codex trimmed the
screenshot turns from its session history, so the games cannot be reconstructed. We therefore report
them as an unverified observation and leave them out of Table~\ref{tab:compare} and of every
statistical statement.

DeepSeek V4.1 Flash 1017, served on our own GPUs, played the same 25 seeds through the same
interface and runner. We stopped this cohort after 12 finished games, all lost, because a win had become unlikely.
Six games stopped mid-run and six unstarted games are left out of the count, and one game ended when the model
reached its 40,000-token output limit three times at the same decision without answering. Its runs ended early:
the mean floor was 12.4, against 26.6 for our released policy on the same 12 seeds (one-sided 95\% upper bound
on its win rate 22\%).
We release every decision of both text-interface cohorts' games with the frozen protocols: the chosen actions,
the model's stated reasons and notes, latencies and token counts. Replaying the recorded actions in our host
reproduces the logged outcome of all 23 finished games. The release leaves out the rendered observations because
they contain the game's own text; replaying a game with a copy of the game regenerates them exactly.

\paragraph{Computation per decision.} We timed the released network in the C\# runtime that the mod
uses, on 500 recorded observations, pinned to one CPU core. A decision takes 21--27~ms on one core of
a Ryzen Threadripper 9980X and 62--67~ms on one core of a Xeon CPU Max 9480; with all threads
available, it takes 10~ms on the Threadripper. For each released policy the C\# outputs match PyTorch
on all 500 observations, with the same greedy action every time and logits within
$7.3\times10^{-5}$. A winning A10 run has a median of about 750 decisions, so it needs about 20 seconds of
single-core inference. Spire Mentor reports 14~ms per 1,000 simulations on one thread of a different
CPU, about 112~ms per combat decision at 8,000 simulations and 224~ms at the 16,000 its newest
Ironclad uses in boss fights. Its Ironclad A10 runs took 162~s of CPU each in its own benchmark on a
shared host. Its search also relies on a complete reimplementation of the game kept in step with the
live game, while our policy needs only the observation. GPT-6 Astra in our text interface took a
median of 13.7~s per decision, with about 10,400 input and 320 output tokens, and the cohort cost
about 70~USD per hour (mix of credits and subscription) at eight concurrent games. DeepSeek V4.1 Flash on our own servers took a median
of 29~s. One network evaluation per decision therefore costs several times less than tree search on
a CPU core and about three orders of magnitude less time than a language model. At that cost it
reaches 8.1\% at A10, far above AlphaSpire's networks used without search and the language models
we ran, and below tree search.

\section{Controlled study of restart allocation}
\label{sec:screen}
\subsection{Design}
We compare three training methods from scratch: fresh starts only, restarts with uniform situation
allocation, and restarts with failure-driven allocation. Both restart methods use the same
within-situation sampling, buffers and restart probability, so the comparison of the two isolates the
outer allocation rule. The study covers Regent and Silent at Ascension~0 with three independent random
initializations per character, giving 18 runs. Within each character--initialization block, the
three methods start from identical weights with a new optimizer. Table~\ref{tab:screen} lists the
frozen design.

\begin{table}[ht]
\centering
\caption{Design of the independent-start study, frozen before training.}
\label{tab:screen}
\begin{tabular}{p{0.30\linewidth}p{0.61\linewidth}}
\toprule
Factor & Setting\\
\midrule
Methods & Fresh only; uniform situations + uncertainty; failure map + uncertainty\\
Task & Regent and Silent, A0, single player, no ascension curriculum\\
Budget & 763 updates of 65,536 rows; 50,003,968 rows per run\\
PPO & learning rate $2\times10^{-4}$; $\gamma=0.9998$; $\lambda=0.99$; two epochs; minibatch 4,096\\
Workers & 256 simulator processes\\
Development curves & EMA at updates 153, 306, 459, 611 and 763 on 256 development games\\
Final test & EMA at update 763 on 1,024 held-out games\\
\bottomrule
\end{tabular}
\end{table}

The budget fixes the number of optimizer rows. Restart runs also replay action prefixes, which costs
simulator time outside this budget (Table~\ref{tab:screen-cost}), so the study compares methods at
equal optimization samples, with total simulator work recorded separately. All arms restart their
worker processes and prefix buffers at the same five segment boundaries, while model, optimizer, EMA,
sampler statistics and learner random state carry over. During validation we found that the first
worker implementation could give consecutive combats the same identifier when no non-combat decision
separated them, which mislabeled the auxiliary HP target and restart points for the two A10 final
bosses. We corrected the worker, verified five complete A10 regression runs with identical actions and
observations, discarded the affected batch of 59.1M rows, and ran all 18 runs anew from the original
six initializations.

\subsection{Results}
All 18 final policies played the same 1,024 held-out games, 18,432 in total, with no engine error.
Table~\ref{tab:screen-results} gives every training block.

\begin{table}[ht]
\centering
\caption{Final A0 wins out of 1,024 common held-out games. Each row is one random initialization
shared by the three methods; bold marks the best method of the row. The last column is the paired
difference in win rate, in percentage points.}
\label{tab:screen-results}
\begin{tabular}{llrrrr}
\toprule
Character & Initialization & Fresh & Uniform & Failure & Failure $-$ uniform\\
\midrule
Regent & 41001 & 74 & 54 & \textbf{99} & \incr{4.39}\\
Regent & 41002 & 60 & 54 & \textbf{94} & \incr{3.91}\\
Regent & 41003 & 80 & 106 & \textbf{131} & \incr{2.44}\\
Silent & 52001 & \textbf{114} & 72 & 86 & \incr{1.37}\\
Silent & 52002 & 28 & \textbf{53} & 40 & \decr{1.27}\\
Silent & 52003 & 29 & 35 & \textbf{46} & \incr{1.07}\\
\bottomrule
\end{tabular}
\end{table}

For Regent, failure-driven allocation wins 10.55\% on average, against 6.97\% for both the fresh
and the uniform arm. It exceeds uniform allocation in all three blocks, by $+3.58$ points on average
(range $+2.44$ to $+4.39$). For Silent, the three arms average 5.57\%, 5.21\% and 5.60\%; the
failure-minus-uniform difference averages $+0.39$ points with blocks of both signs. Variation between
initializations is large for Silent, where fresh training ranges from 28 to 114 wins. Three blocks per
character show a consistent direction for Regent and no effect for Silent. They do not establish the
size of the Regent effect for other initializations, other characters or higher difficulty. These
short-budget policies win far less than the released ones, which collected at least twenty times
more decisions.

\begin{figure}[ht]
\centering
\includegraphics[width=\linewidth]{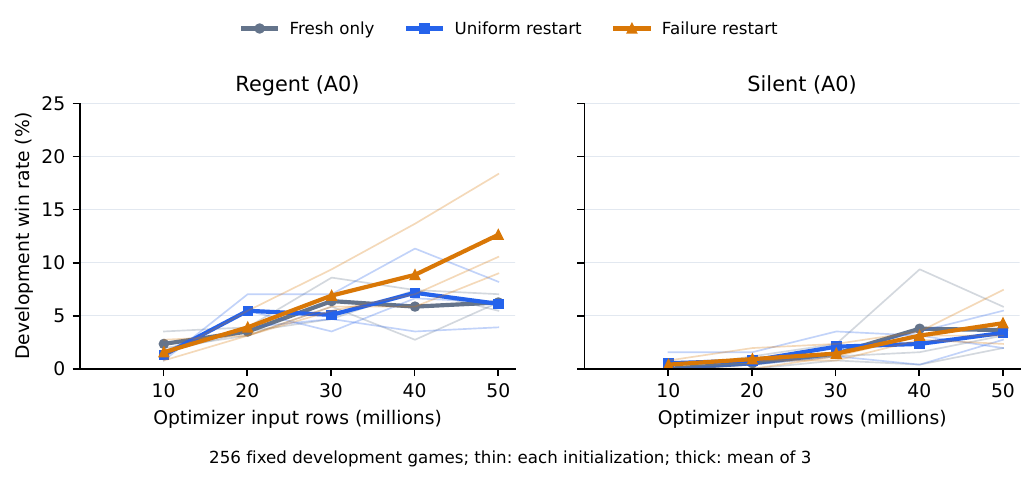}
\caption{Development curves for all 18 runs. Thin lines are single initializations and thick lines
the means of three; each point uses 256 development games. Development games are separate from the
final test and were not used to choose checkpoints.}
\label{fig:screen-learning}
\end{figure}

\begin{figure}[ht]
\centering
\includegraphics[width=.94\linewidth]{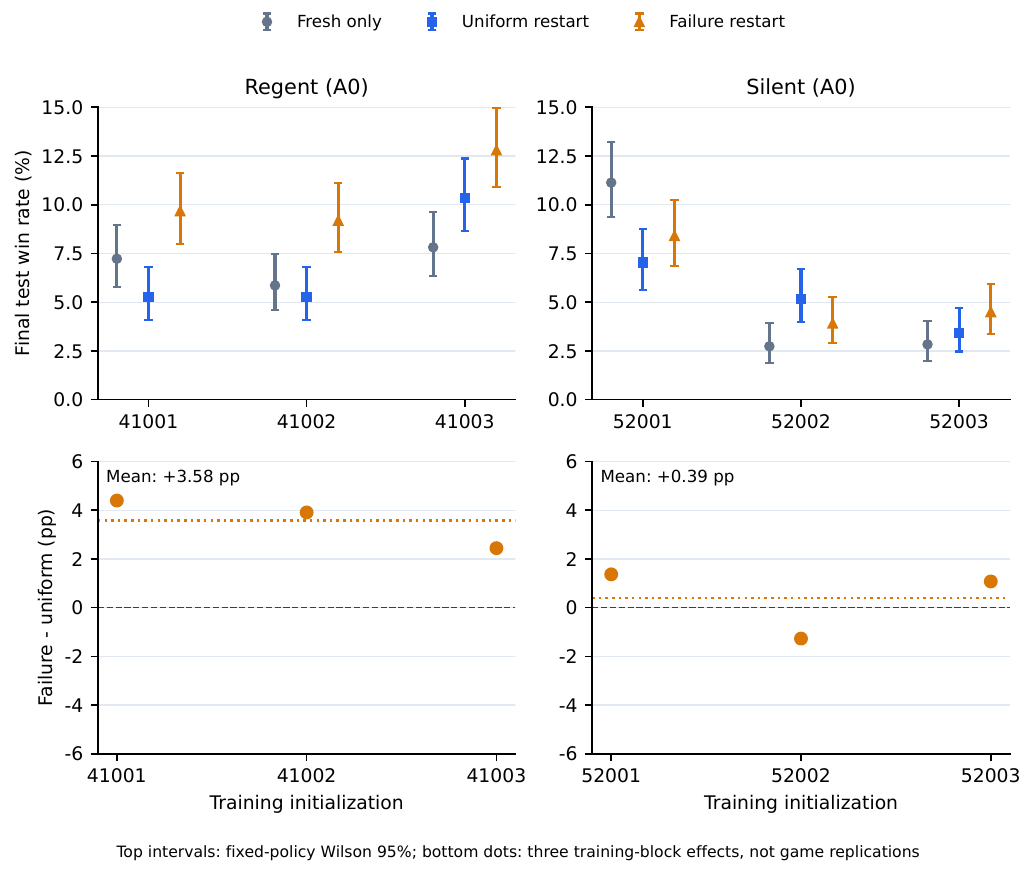}
\caption{Top: final policies with Wilson 95\% intervals over test games. Bottom: the three
independent block differences per character and their means.}
\label{fig:screen-final}
\end{figure}

\begin{table}[ht]
\centering\small
\caption{Recorded training work, summed over the six runs of each method. Decisions in millions.
Trainer hours are summed over segments on mixed hardware and do not form a speed comparison.}
\label{tab:screen-cost}
\begin{tabular}{lrrr}
\toprule
Quantity & Fresh & Uniform & Failure\\
\midrule
Optimizer rows (M) & 300.024 & 300.024 & 300.024\\
Collected decisions (M) & 300.638 & 300.618 & 300.621\\
Replayed prefix decisions (M) & 0 & 149.819 & 134.631\\
Restart episodes & 0 & 591,062 & 577,858\\
Completed episodes & 895,131 & 1,187,028 & 1,161,006\\
Summed trainer hours & 15.17 & 17.41 & 17.88\\
\bottomrule
\end{tabular}
\end{table}

\begin{figure}[ht]
\centering
\includegraphics[width=\linewidth]{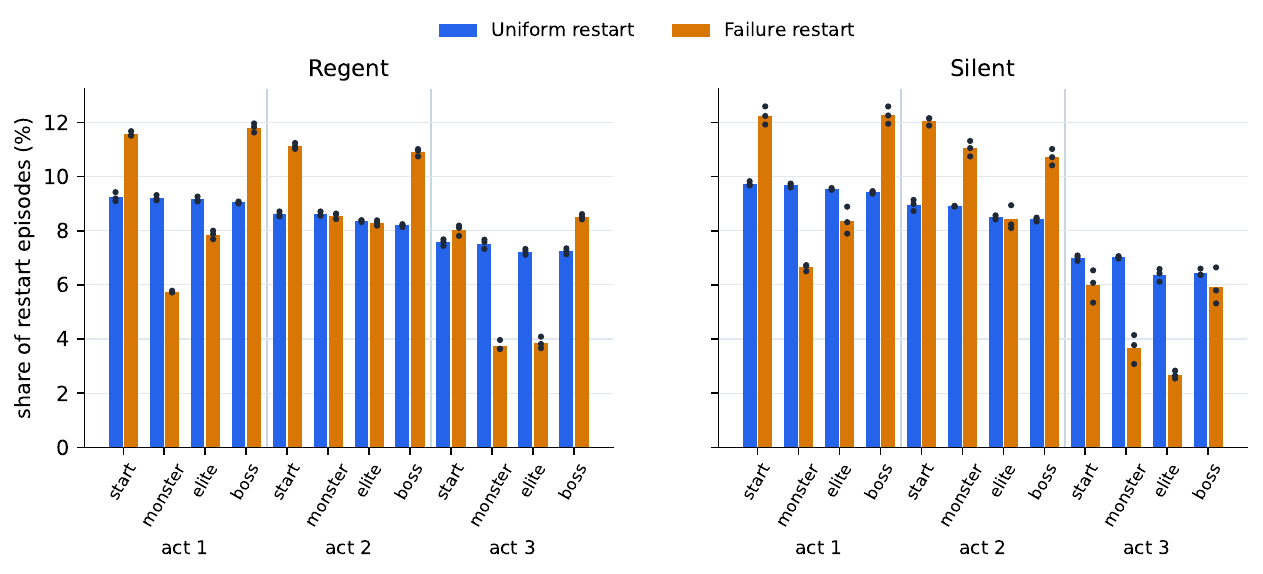}
\caption{Share of restart episodes by starting situation over the whole run. Bars are means over the
three initializations; dots are single runs.}
\label{fig:alloc}
\end{figure}

Figure~\ref{fig:alloc} shows where restart episodes started. Uniform allocation gives each
situation a similar share; later situations receive somewhat less because workers hold stored states
for them less often. Failure-driven allocation moves practice toward act starts and the act-1 and
act-2 bosses and away from act-3 monster and elite fights, in a similar pattern for both characters.
One visible difference is the act-3 boss, which failure allocation practices more than uniform
allocation for Regent (8.5\% against 7.2\%) and less for Silent (5.9\% against 6.4\%). This describes
three runs per arm and does not establish a cause of the different outcomes.

Restart arms replay about half as many prefix decisions as they collect. Requested and actual restart
counts are equal in every run because of the availability masking. Failure-driven allocation replays
10\% fewer prefix decisions than uniform allocation while winning more for Regent, so its Regent gain
does not come from extra simulator work.

\section{Improving the A10 policies}
\label{sec:improve}
We made several attempts to raise the base A10 policies, each with a frozen protocol and fresh test
seeds. Only longer A10 training produced confirmed gains, and its last run gave the released Ironclad,
Defect and Silent policies.

\paragraph{Checkpoint selection.} A fixed pool of 21 historical checkpoints, including the five
base ones, played 256 development games per character at A10. The best candidate per character
then played 1,024 new test games against the base policy. The five-character mean moved from
6.41\% to 6.66\%, with a 95\% bootstrap interval over common seeds of $-0.53$ to $+1.04$ points, and
no paired comparison approached significance (Table~\ref{tab:historical}).

\begin{table}[ht]
\centering\small
\caption{Checkpoint selection at A10, wins out of 1,024 test games; bold marks the higher count.
Paired exact $p$-values; all equal 1 after Bonferroni correction over five characters.}
\label{tab:historical}
\begin{tabular}{lrrrr}
\toprule
Character & Base & Selected & Difference (pp) & $p$\\
\midrule
Ironclad & 122 & \textbf{124} & \incr{0.20} & .939\\
Silent & 20 & \textbf{22} & \incr{0.20} & .860\\
Defect & 36 & \textbf{39} & \incr{0.29} & .795\\
Necrobinder & 15 & 15 & 0.00 & 1.000\\
Regent & 135 & \textbf{141} & \incr{0.59} & .705\\
\bottomrule
\end{tabular}
\end{table}

\paragraph{Short continuation.} Each base policy trained for 5.05M more rows at A10 with the
failure-driven restart curriculum and a learning rate of $3\times10^{-5}$. A candidate replaced the
base policy only if it won strictly more of 256 development games. On 1,024 test games per
character the mean moved from 6.09\% to 6.45\% (bootstrap interval $-0.31$ to $+1.02$ points;
Table~\ref{tab:continuation}). Regent's gain of 2.93 points had an unadjusted paired $p=0.018$ and
$p=0.088$ after correction for five comparisons. A separately frozen confirmation on 2,048 new seeds
gave 251 wins for the base Regent policy and 252 for the candidate (153 games gained, 152 lost,
exact $p=1.0$), so the gain did not replicate. Ironclad's development gain from 26 to 41 wins reversed
on the test panel.

\begin{table}[ht]
\centering\small
\caption{Short A10 continuation. Development: base $\to$ candidate, wins of 256. Test: base
$\to$ selected, wins of 1,024. Silent and Defect kept the base policy. Bold marks the higher count.}
\label{tab:continuation}
\begin{tabular}{lrrr}
\toprule
Character & Development & Test & Difference (pp)\\
\midrule
Silent & $\mathbf{4}\to1$ & $19\to19$ & 0.00\\
Defect & $\mathbf{9}\to4$ & $29\to29$ & 0.00\\
Necrobinder & $3\to\mathbf{5}$ & $\mathbf{18}\to13$ & \decr{0.49}\\
Ironclad & $26\to\mathbf{41}$ & $\mathbf{128}\to121$ & \decr{0.68}\\
Regent & $32\to\mathbf{35}$ & $118\to\mathbf{148}$ & \incr{2.93}\\
\bottomrule
\end{tabular}
\end{table}

\paragraph{Longer continuation.} Ironclad and Regent continued from their base weights for 600
updates (39.3M rows) at A10 with the recipe that ended their development history. On a 1,024-seed
panel the Ironclad candidate won 138 games against the base policy's 99 (paired $p=.003$), a
comparison outside the frozen readout of that experiment. A confirmation frozen before any game, on
2,048 new seeds, gave 268 against 247 for Ironclad ($p=.28$) and 305 against 266 for Regent ($p=.047$,
Holm-adjusted $.093$). Neither passes; the first panel happened to be one on which the base Ironclad
policy did poorly (9.7\% against 12.1\% on the confirmation seeds). We do not pool the panels.

\paragraph{Second-boss practice.} A curriculum aimed at the final double boss restarted training
episodes from second-boss situations stratified by entry HP, added a per-fight outcome reward and
repeated each practiced fight four times. Against the plain continuation above, with the same
updates, seeds and 1,024 test games, Ironclad was unchanged (141 against 138, $p=.88$) and Regent
fell (102 against 138; 65 against 101 discordant, Holm-adjusted $p=.013$).

\paragraph{Search at the final bosses, and imitating it.} We replayed our policy's own runs to the
first final boss and let Spire Mentor's search (8,000 simulations) play both final bosses
from those entries: 251 Ironclad and 170 Regent entries from 400 seeds each. The search won 66 and 72
runs where our policy won 51 and 61 (pooled, 52 runs won only by search against 26 only by our
policy, $p=.004$), and it won the second boss more often from identical boss-2 states (19 against 7
and 20 against 5). Searching only the final bosses would thus raise these A10 rates by about three
points (3.8 for Ironclad, 2.8 for Regent). We then distilled the search's choices into the network,
training on our policy's own boss states with a target of 0.8 on the search's pick plus 0.2 on its
visit shares and a KL anchor to the base policy on full-run states. Agreement with the search's pick on held-out states rose from
45.7\% to 50.0\% (Ironclad) and from 40.1\% to 45.6\% (Regent), while the greedy move changed in 31\%
and 41\% of decisions. From the same boss-1 entries the distilled policies won 60 and 62 runs against
74 and 74 for the base ones (pooled 29 against 55, $p=.006$). An earlier variant trained on visit
shares alone also lost (63 against 74, Ironclad). The search's visit distributions are nearly flat,
and the network could not predict its picks from its own observation even where the search was
confident, so imitation added noise to otherwise coherent play.

\paragraph{Long A10 training for the weakest characters.} The A10 recipe that raised Ironclad and
Regent during development combined single-character training at A10, the failure-driven curriculum
and a learning rate of $10^{-4}$, over several hundred million decisions. Silent, Defect and
Necrobinder never received it. We trained each of them from its base weights for 201.3M
optimizer rows (3,072 updates) at A10 with this recipe. The frozen protocol names the final EMA as
the only candidate and compares it with the base policy on 1,024 new test seeds per character,
with an exact McNemar test and Holm correction over the three characters; it also checks A0 and A9 on
the 512 sweep seeds. Defect rose from 31 to 80 wins (3.0\% to 7.8\%; 71 seeds won
only by the new policy against 22, Holm-adjusted $p<10^{-5}$) and Silent from 16 to 31 (1.6\% to
3.0\%; 27 against 12, $p=.047$); Necrobinder's rise from 13 to 20 is not significant ($p=.28$)
(Table~\ref{tab:push}). All three reach 6 to 8 floors deeper on average. At A9 the new policies win
more often for all three characters, clearly so for Defect (125 against 81 of 512, $p<.001$). Across
difficulty, the trained Defect policy also wins more often at A4 (393 against 358 of
512, $p=.009$) and A8 (282 against 235, $p<.001$), while the trained Silent and Necrobinder policies are level
with the base ones at A4 and A8 (all $p>.3$). At A0
Silent and Defect hold their rates, while Necrobinder falls from 372 to 345 of 512 ($p=.044$,
uncorrected), so for that character long A10 training traded some low-difficulty play for A10 and A9
play. Defect's gain came from a recipe that the base Ironclad and Regent policies had already
received.

A second round of the same recipe for Defect, frozen before training and started from the first round's
final EMA (another 3,072 updates), did not raise it further. On a second set of 1,024 new A10 test seeds
it won 88 games against 82 for the first-round policy (8.6\% against 8.0\%; 55 seeds won only by the
second round against 49, $p=.62$). At A0 and A9 on the 512 sweep seeds it won 422 against 400
($p=.059$) and 135 against 125 ($p=.45$). The first-round policy's 8.0\% on these fresh seeds
replicates its 7.8\% on the first test panel.

\paragraph{Release continuation.} The last run of this recipe produced the released Ironclad, Defect
and Silent policies. Each continued from its base weights for 1,800 updates (118M rows) at A10 with the
failure-driven curriculum and a learning rate of $10^{-4}$, as the control arm of a frozen 12-run study
whose other arms we will report separately; the protocol named the final EMA as the only candidate. On
that study's 2,048 new A10 test seeds per character, the continued policies won 313, 176 and 54 games
against 234, 53 and 33 for the base policies (seeds won only by the continued / only by the base
policy: Ironclad 216 / 137, $p<10^{-4}$; Defect 158 / 35, $p<10^{-19}$; Silent 44 / 23, $p=.014$). These
comparisons lie outside the study's frozen readout, which compares training methods with this control,
and an independent replicate of the same arm reached 330 Ironclad and 156 Defect wins. We adopted the
continued policies as the release after these results; Table~\ref{tab:a10} then measured them, under a
plan frozen before any game, on 2,560 held-out games per character that took no part in that choice.
There Ironclad won 399 games against 317 for its base policy (269 seeds won only by the continued
policy against 187, $p<10^{-3}$), Defect 208 against 78 (186 against 56, $p<10^{-16}$) and Silent 61
against 47 (54 against 40, $p=.18$): the gains confirm for Ironclad and Defect, and Silent's remains
uncertain. This run is three
times as long as the longer continuation above (1,800 against 600 updates), which did not confirm for
Ironclad.

\begin{table}[ht]
\centering\small
\caption{Long A10 training (201.3M rows) for the three weakest characters. Wins of 1,024 new A10
test seeds; ``discordant'' counts the seeds won only by the trained / only by the base policy
(exact McNemar, Holm over three characters). Mean floor reached, and wins of 512 on the A0 and A9
sweep panels, base $\to$ trained; bold marks the better value. The released Silent and Defect
policies come from the release continuation instead.}
\label{tab:push}
\setlength{\tabcolsep}{4pt}
\begin{tabular}{lrrrrrrr}
\toprule
Character & Base & Trained & Discordant & Holm $p$ & Floor & A0 & A9\\
\midrule
Silent & 16 & \textbf{31} & 27 / 12 & .047 & 24.7$\to$\textbf{32.3} & 335$\to$\textbf{345} & 53$\to$\textbf{66}\\
Defect & 31 & \textbf{80} & 71 / 22 & $<10^{-5}$ & 32.4$\to$\textbf{38.8} & \textbf{407}$\to$400 & 81$\to$\textbf{125}\\
Necrobinder & 13 & \textbf{20} & 19 / 12 & .28 & 24.1$\to$\textbf{30.0} & \textbf{372}$\to$345 & 30$\to$\textbf{46}\\
\bottomrule
\end{tabular}
\end{table}

\section{Discussion and limitations}
A single network evaluation per decision carries the released policies at A10 to about one win in six
for Ironclad, one in eight for Regent and one in twelve for Defect, and to one in forty or fewer for
Silent and Necrobinder. In our host tree search leads clearly for Silent, leads without a significant
difference for Ironclad and Defect, is level for Necrobinder and is behind for Regent. Search can
correct a combat line at the moment of play, and the final double boss
rewards exact combat sequencing over many turns, so this split is consistent with where our policies
lose. The comparison has limits. Mentor's reported numbers come from its own engine, seed panels and
observation contract; our in-host run removes the engine and panel differences but not every detail
of its real-game driver. Our observation includes the enemies' internal next-move
identifier. The host patches parts of the game for headless execution, and behavioral equivalence with
the live game is tested on selected cases only.

Why do A10 rates move so little? Three measurements point the same way. First, the win is a product
of stages: in our replay of the base Ironclad policy on 400 seeds, 63\% of runs reach the first final boss, 64\% of those
win it and 32\% of those win the second, so a five-point gain at either of the first two stages moves
the run rate by about one point (two points at the second boss), and a one-point change takes thousands
of paired games to detect. Second, the decisive errors come
early: for 90 of 125 final-boss losses (72\%), none of 12 replays that re-decided all of act~3 won, and
for 72 of them (58\%) no replay of the boss fights with the same dice won either, so the deck built in
acts 1 and 2 could not take both bosses. Third, the policy barely conditions on
difficulty. On 139,744 decisions with more than one legal action from its own A10 runs, setting the
ascension input from A10 to A0 changes the base Ironclad policy's greedy action in 5.9\% of them (Regent 3.4\%; shop decisions change
most, 15.0\% for Ironclad and 10.5\% for Regent), while its win-probability head moves from 0.16 to 0.80 (Regent 0.24 to 0.70).
The win-probability head separates the levels, and the policy hardly does. The development history trained mostly
below A10, through an ascension curriculum, with A10-only training at the end for Ironclad and Regent
and none for the other three, so a strategy that wins nine runs in ten at A0 may be reused almost
unchanged where it wins one in eight.

The restart study isolates one design choice and shows a character-dependent effect with three
initializations per character. It equalizes optimizer rows and records the extra prefix replay
separately. It does not compare against faithful implementations of related curricula, and it was run
at A0 with short budgets. The base policies come from an exploratory development history whose
individual steps were not run as controlled experiments; only the final continuation of three of them
ran under a frozen protocol.

\section{Conclusion}
Running the game's own code headlessly makes exact replay available, and exact replay makes restart
curricula practical in a long-horizon card game. Zero-search policies trained this way reach 8.1\% at
STS2 A10 over five characters, 15.6\% for Ironclad and 12.5\% for Regent, the highest rates we know of
for agents that use one network evaluation per decision. Failure-driven restart allocation helped
Regent consistently in a controlled study and did not help Silent. Closing the gap to search at A10 will likely require stronger combat play in the final fights, for
example by distilling search into the policy. Code, protocols, evaluation records and the five
policies are released (Appendix~\ref{app:repro}).

\appendix
\section{Reproduction and artifacts}
\label{app:repro}
The release has three parts. The repository \url{https://github.com/HalfVulpes/sts2ai} holds the
source code: the headless host, the trainer with its restart curricula, the in-game mod and the tools
below. Its release page holds the playable mod with the five released policies, each policy's exported
network for quick evaluation, and their PyTorch weights. A records archive,
\url{https://drive.google.com/drive/folders/1OP9IzFIYPuyR97eg1XZru6bpZDSh87Oz}, linked from the
repository, holds the frozen protocols, per-game results, the training logs of the released
runs, the text-interface game logs and the records behind the Discussion.
\begin{enumerate}
\item Import a local v0.107.1 installation with \texttt{tools/import\_game.py}, which verifies all
  eleven assembly hashes, then build the host and generate the vocabulary locally. The release
  contains no game files.
\item Run the source tests without game files; integration tests additionally need the local game.
\item Each study has a frozen protocol in the records archive's \path{protocols/} folder, with its
  seed panels, initial tensor hashes and decision rules. Per-game results, commands, hardware records
  and model hashes are in \path{records/}, and training update logs in \path{training_logs/}; the
  archive's index maps every table and figure to its files.
\item Evaluate a policy greedily with \texttt{scripts/evaluate.py} or the host's \texttt{play}
  command, which take the exported policy, character, ascension and canonical seed strings; every
  game counts, and engine errors are reported. \texttt{scripts/train.py} prints and runs the trainer
  command of each reported recipe.
\end{enumerate}
Original code is released under GPL-2.0-only. The repository holds the algorithms and methods needed
to train the policies from scratch, including the restart curricula of the longer continuation,
second-boss practice and the release continuation; it leaves out platform-specific job tooling. The
Spire Mentor adapter of our host is not part of the release; the protocols and per-game records of
the studies that used it are in the archive.

\section{Hyperparameters}
\begin{table}[ht]
\centering\small
\caption{PPO and curriculum settings shared by all runs in this paper unless a study states otherwise.}
\begin{tabular}{llll}
\toprule
Parameter & Value & Parameter & Value\\
\midrule
$\gamma$ / GAE $\lambda$ & 0.9998 / 0.99 & samples per iteration & 65,536\\
clip (policy / value) & 0.2 / 0.2 & epochs, minibatch & 2, 4,096\\
weight decay & 0.01 & gradient-norm clip & 1.0\\
entropy (combat / other) & 0.003 / 0.01 & value / auxiliary coefficient & 0.5 / 0.2\\
EMA decay per step & 0.998 & restart probability & 0.5\\
situations & 12 & stored states per situation & 16 per worker\\
failure-map round, decay & 10 iterations, 0.5 & combat credit, floor & 0.65, 2\%\\
\bottomrule
\end{tabular}
\end{table}

\section{Continuation pilot}
Before the independent-start study, nine short runs continued one historical Regent checkpoint for
about two million decisions each, three per method with three continuation seeds. On the same 200
legacy development seeds the initial policy won 140. Table~\ref{tab:pilot} lists every run. Failure
minus uniform was $+6.0$, $-3.5$ and $+4.0$ points; all nine runs share one pretrained source, so
the pilot measures adaptation of one policy.

\begin{table}[ht]
\centering
\caption{Regent A0 continuation pilot, wins of 200 common legacy development games.}
\label{tab:pilot}
\begin{tabular}{lrrrr}
\toprule
Continuation seed & Fresh & Uniform & Failure & Failure $-$ uniform\\
\midrule
101 & 151 & 141 & \textbf{153} & \incr{6.0} pp\\
202 & 132 & \textbf{144} & 137 & \decr{3.5} pp\\
303 & 130 & 126 & \textbf{134} & \incr{4.0} pp\\
\bottomrule
\end{tabular}
\end{table}

\section{Text interface and earlier language-model diagnostics}
The text interface runs the same host and exposes three operations: read the state, choose a numbered
action and store a note. It offers no reset, lookahead or hidden state, and every action is logged.
A known rendering defect prints some energy icons as resource paths inside card text. Before the
Astra cohort, GPT-6.1 Terra at high reasoning (one level below maximum) played one game per character
under the same bounded-context protocol and lost all five, ending on floors 17, 42, 12, 17 and 14; the
released RL policies won one of the same five seeds. An earlier single Ironclad game switched
controllers midway and is excluded from all comparisons.

\section{Legacy development evaluations}
Two development evaluations of the released policies on legacy numeric seeds, 50 games per character
and level at A0, A9 and A10, gave a five-character mean of 4.0\% at A10 under both GPU and C\#
inference, with A0 between 66\% and 92\% per character. Greedy actions matched across the two backends
on 32 recorded observations per model, and four complete Ironclad A10 runs matched in outcome, floor
and decision count. These numbers use a different seed distribution and are not pooled with the
canonical panels.

\end{document}